\documentclass{article}

    \PassOptionsToPackage{numbers, compress}{natbib}

    \usepackage[preprint]{neurips_2021}

\usepackage[utf8]{inputenc} 
\usepackage[T1]{fontenc}    
\usepackage{hyperref}       
\usepackage{url}            
\usepackage{booktabs}       
\usepackage{amsfonts}       
\usepackage{nicefrac}       
\usepackage{microtype}      
\usepackage{xcolor}         
\usepackage[utf8]{inputenc}
\usepackage[small]{caption}
\usepackage{booktabs}
\usepackage{diagbox}
\usepackage[flushleft]{threeparttable}
\usepackage{times}
\usepackage{epsfig}
\usepackage{graphicx}
\usepackage{amsmath}
\usepackage{amssymb}
\usepackage{mathtools}
\usepackage{algorithm}
\usepackage{algpseudocode}
\title{Multimodal Shape Completion via IMLE}

\author{%
  
  Himanshu Arora \\
  Simon Fraser University \\
   \texttt{haa48@sfu.ca} \\
   
   \And
   
   Saurabh Mishra \\
   Flipkart and Simon Fraser University\\
   \texttt{saurabh\_mishra@sfu.ca}
   
   \And
   
   Shichong Peng \\
   Simon Fraser University \\
   \texttt{shichong\_peng@sfu.ca} \\
   
   \And 
   
   Ke Li \\
   Simon Fraser University \\
   \texttt{keli@sfu.ca} \\
   
   \And
   
   Ali Mahdavi-Amiri\\
   Simon Fraser University \\
   \texttt{amahdavi@sfu.ca}
}

\usepackage{color}
\definecolor{green}{rgb}{0, 0.5, 0}
\definecolor{orange}{rgb}{0.8, 0.6, 0.2}
\definecolor{red}{rgb}{1.0, 0.0, 0.0}
\definecolor{teal}{rgb}{0.0, 0.4, 0.4}
\definecolor{purple}{rgb}{0.65,0,0.65}
\definecolor{saffron}{rgb}{0.95,0.75,0.2}
\definecolor{turquoise}{rgb}{0.0,0.5,0.5}
\definecolor{black}{rgb}{0.0, 0.0, 0.0}
\definecolor{gray}{rgb}{0.5, 0.5, 0.5}

\newcommand{\am}[1]{{\color{black}#1}}

\newcommand{\nv}[1]{\mathbf{#1}}

\begin{document}

\maketitle

\begin{center}
    \includegraphics[width=1\textwidth]{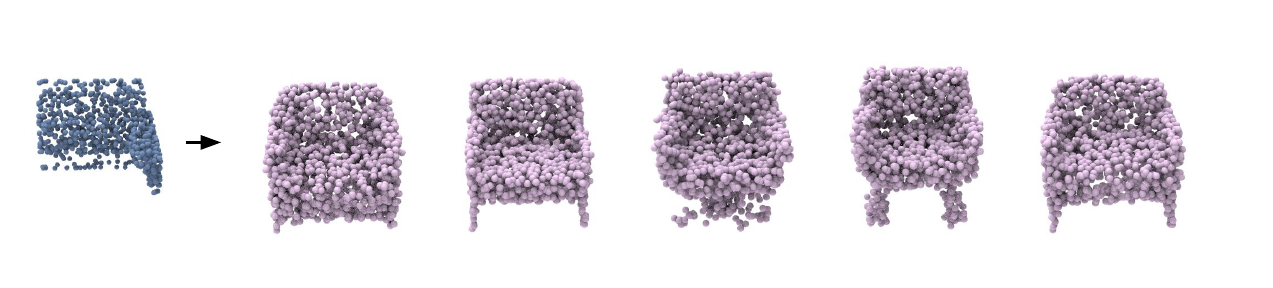}
    \captionof{figure}{Having a partial point cloud as input, our method is capable of producing \emph{diverse} complete outputs. }
    \label{fig:teaser}
\end{center}%

\begin{abstract}
 Shape completion is the problem of completing partial input shapes such as partial scans. This problem finds important applications in computer vision and robotics due to issues such as occlusion or sparsity in real-world data. However, most of the existing research related to shape completion has been focused on completing shapes by learning a one-to-one mapping which limits the \emph{diversity} and \emph{creativity} of the produced results. 
 We propose a novel multimodal shape completion technique that is effectively able to learn a one-to-many mapping and generates diverse complete shapes. 
 Our approach is based on the conditional Implicit Maximum Likelihood Estimation (IMLE) technique wherein we condition our inputs on partial 3D point clouds. We extensively evaluate our approach by comparing it to various baselines both quantitatively and qualitatively. We show that our method is superior to alternatives in terms of completeness and diversity of shapes.
\end{abstract}

\section{Introduction}
\label{sec:intro}

\am{As 3D scanning devices are now widely available, scenes and objects in the form of point clouds are rapidly becoming more and more prevalent. 
However, many such objects are not fully scanned due to occlusion; as a result, the scanned shapes may be missing substantial portions of the complete shape. 
Therefore, the ability to complete such shapes by generating the missing portions is desired. 
More precisely, the objective of shape completion is to meaningfully complete a partial shape and obtain a complete shape without violating the geometry of the partial shape.}

\am{Early work has focused on generating a single complete shape; the mapping from the input and output is one-to-one.
However, shape completion is ill-posed by nature as the target shape is not unique -- given a partial shape, there are multiple ways to generate the missing portions, especially when the input point cloud is noisy and coarse, which is the case in real-world scanned data. 
Therefore, given a partial shape, it is desired to generate a \emph{set} of many possible complete shapes, all of which respecting the given input partial shape. 
Not only is multimodality a way to resolve ill-posedness, but it also adds more \emph{diversity} and \emph{creativity} to the final outputs.}

\begin{figure}
    \includegraphics[width=1\linewidth]{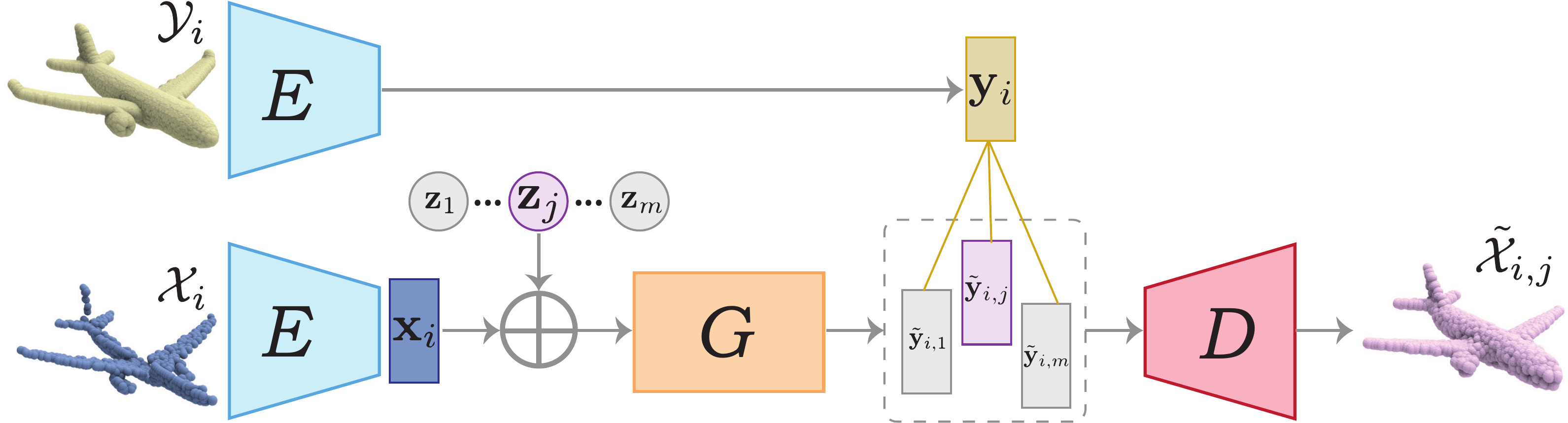}
    \caption{Our network receives a partial point cloud $\mathcal{X}_i$ and a complete point cloud (in training time) $\mathcal{Y}_i$. The latent code generated from the incomplete shape, $\nv{x}_i$ along with a noise, $\nv{z}_j$ is given to generator $G$ to be mapped to $\nv{\tilde{y}}_{i,j}$. Closest $\nv{\tilde{y}}_{i,j}$ to the latent code $\nv{y}_i$ of the complete point cloud is found and passed to decoder $D$ to produce a novel and complete shape $\mathcal{\tilde{X}}_{i,j}$. }
    \label{fig:pipe-line}
\end{figure}

\am{Recently, Wu et al. \cite{wu2020multimodal} proposed the task of multimodal shape completion, where the goal is to generate various complete shapes for a single partial shape. This is challenging, because the training data only contains one ground truth complete shape for each partial shape. 
In \cite{wu2020multimodal}, a conditional GAN-based model was proposed to learn a one-to-many mapping from partial shapes to complete shapes. 
}


While \cite{wu2020multimodal} took a step towards multimodal shape completion, the different shapes it generates for a given partial point cloud tend to be similar in structure. For example, for an incomplete chair, the different complete planes it generates are of a similar style and vary mostly in width and height, as shown in Figure \ref{fig:vis3}. In this paper, our goal is to generate \emph{structurally diverse} complete shapes for the same incomplete shape. In particular, consider the case where a part of the object is entirely missing, the different generated complete shapes should ideally vary in the style of the missing part while remaining compatible with the other parts that are present.

Our key insight is that structural similarity between the generated complete shapes comes from the tendency of GANs to collapse modes. To achieve structural diversity, we must therefore get around mode collapse. To this end, we propose a novel method by devising a model formulation that allows for training with the method of Implicit Maximum Likelihood Estimation (IMLE). Prior to our work, IMLE has never been successfully used for point cloud generation. In this paper, we show how to do so, which results in a simple and elegant model with fewer parameters than \cite{wu2020multimodal}. Yet, despite this simplicity, we demonstrate that our method is capable of generating structurally diverse results. Quantitatively, our method achieves superior performance than prior methods in terms of Total Mutual Difference, and comparable or superior performance than prior methods in terms of Unidirectional Haussdorf Distance (UHD). Moreover, we demonstrate the capability of our method to generalize to new kinds of incomplete shapes. Despite not having part-level supervision, our method can produce complete shapes for incomplete shapes with an entire part missing.


\section{Related Work}
\label{sec:RW}

\am{Since the missing regions of an incomplete point cloud are unknown and might be geometrically complex, simple surface repair algorithms are not fully effective in reproducing parts and geometric features. In fact, missing regions should be better \emph{generated} after observing many relevant shapes to learn their geometric properties.  Therefore, here, we mostly provide a review of deep generative models that are reasonable solutions for shape completion.
In addition, since our input is a point cloud, we provide a review on deep learning models designed for point clouds, and we particularly discuss point cloud completion methods.
Point cloud completion may have similarities with other problems, including point upsampling, 3D part assembly, 3D reconstruction, and single view reconstruction. However, since the focus of our paper is multimodal point cloud completion, we refrain from discussing works related to those areas and refer the interested reader to the available survey papers such as \cite{egstar2020_struct,berger2014state}.
}

\subsection{Generative Models}
\am{Generating new images or 3D models is an important area in computer vision and computer graphics.
One of the most well-known techniques to generate new shapes is \emph{variational autoencoder} \cite{kingma2013auto} or VAE that tries to fit a known distribution (e.g., Gaussian) to the latent space of an autoencoder so that by sampling the known distribution, the decoder can produce novel data.
VAEs have been used in many applications, including generating new 3D datasets and shape completion \cite{stutz2020learning,egstar2020_struct}. 

\emph{Generative Adversarial Networks} (GANs) \cite{goodfellow2014generative} are alternative generative models composed of a generator and a discriminator, where the generator tries to produce novel data that can fool the discriminator, and the discriminator tries to distinguish the produced results (i.e., fake) from real samples.
There are various forms of GAN that have been designed to tackle different problems, including conditional GANs that try to generate data based on a condition such as an incomplete image for the purpose of inpainting \cite{yu2018generative,isola2017image} or a partial shape for shape completion \cite{chen2019unpaired}.
In fact, GAN-based networks have been successful in many 3D tasks and applications \cite{egstar2020_struct}.
}

\am{Despite the popularity and success of GANs, they have limitations including mode collapse. 
This problem is more evident when variety in the generated data is required (such as the case of multimodal shape completion). 
This has been the primary motivation behind IMLE, which is to avoid mode collapse and it has been applied to various multimodal image generation problems including image super-resolution \cite{Li2020MultimodalIS}, image synthesis from semantic layout \cite{Li2019DiverseIS} and image decompression \cite{Peng2020GeneratingUA}.
}

\subsection{Deep Learning on Point Clouds}
\am{One of the possibilities is to feed a neural network by an array of points (i.e., point clouds) capturing the geometry of shapes.
One of the earliest networks to deal with point clouds is PointNet \cite{qi2017pointnet} that tries to learn feature points to perform different tasks including segmentation or classification.
As PointNet only learns global features, PointNet++  \cite{qi2017pointnet++} is introduced to learn multi-scale features and better abstract local patterns.
Due to the effectiveness of PointNet and PointNet++, they have been used as the main component of many networks with various applications including point to point translation \cite{yin-sig18,yin-siga19}, edge detection on point clouds \cite{wang2020}, learning implicit surfaces \cite{erler2020}, denoising \cite{rakotosaona2020} and many more \cite{guo2020}. }

\subsection{Shape Completion}
\am{\textbf{Classical methods.} Shape completion has been studied in the area of mesh processing even before the popularity of deep learning. 
For example, the missing region of a mesh can be repaired by minimizing an energy function via Laplacian or Poisson equation \cite{sorkine2004least,zhao2007robust,nealen2006laplacian,kazhdan2013screened,kazhdan2006poisson}, benefiting from shapes' skeletal structures \cite{Dpoints15,yin2014morfit}, or retrieving a template from a database and deforming it into the incomplete input point cloud \cite{pauly2005example,rock2015completing}.
Interested readers can refer to \cite{berger2014state} for a thorough study on shape reconstruction and completion for point clouds.}

\am{

\textbf{Deep shape completion.} 
The missing region of a 3D shape, however, might be geometrically complex that can be only \emph{generated} through a deep network trained on a dataset with variety.
As a result, deep learning techniques have been proposed to leverage available 3D shape datasets to learn how to reproduce the missing regions successfully.

In \cite{dai2017shape}, an autoencoder structure is used to first predict a complete distance field of the voxelized incomplete input shape. The distance field is then contrasted against a shape database to obtain the final shape through a multi-resolution patch-based 3D shape synthesis. 
Stutz and Geiger \cite{stutz2020learning} provide a weakly supervised VAE-based method for shape completion. They first train a denoising variational autoencoder on ShapeNet in a supervised manner. They then fix and reuse the decoder to train a new encoder on incomplete shapes where alignment of the predicted result with the incomplete input shape is respected by applying a maximum likelihood (ML) loss.
ASFM-Net \cite{xia2021asfm} maps the partial and complete input point clouds into a common latent space to capture detailed shape priors and better respect the input.

A series of methods have utilized a multi-stage approach to complete a shape.
Inspired from patch-match technique in image processing, Han et al. \cite{han2017high} employed two networks to produce missing parts of an incomplete shape, where one network tries to respect the overall structure of the shape and the other one produces local patches to repair the given incomplete shape. 
PCN \cite{yuan2018pcn} is a multi-stage encoder-decoder network where the decoder first produces a coarse complete shape, and then for each point of the coarse shape, a local patch is generated and deformed to reproduce details. 
Liu et al. \cite{liu2020morphing} also introduced a multi-stage method where a coarse shape is first approximated by a set of parametric surfaces and then it is refined by learning point-wise residuals.
Similarly, Wang et al. \cite{wang2020cascaded} designed a network that performs the shape completion task in two stages (i.e., coarse and dense reconstruction). They have also designed a patch discriminator to guarantee that local patches resemble the same pattern as those in the ground truth.
In addition to the coarse/fine point completion techniques, TopNet \cite{tchapmi2019topnet} has been designed as a tree structure where each branch represents a subset of the point cloud.

\textbf{Detailed shape completion.}
There are recent works that have focused on completing partial input shapes that attain a great amount of detail. 
These works try to replicate existing features in the partial shape in the proper location of the generated complete shape. 
Wen et al. \cite{wen2020point} have defined a skip-attention mechanism to avoid using a global representation that may suffer from altering the initial geometry of the input shape.  
PF-Net \cite{huang2020pf} employs a multi-resolution encoding for the shape and tries to only generate the missing parts of the partial shape to respect the input's features. 
Most recently, VRCNet \cite{pan2021variational} benefits from the multi-stage completion (i.e., coarse and fine stages) and is capable of replicating fine details when they are available. 
Although these works produce nice results when the inputs are detailed, our work is naturally different from these methods. 
Here, we focus on completing input shapes that are relatively coarse and therefore, multiple interpretations are possible.
Only under this setting, the multimodality is meaningful since when the partial input shape is dense and detailed, the complete shape will be limited to only one ground truth shape with the same features as the input. 
In Section \ref{sec:experiments} and supplementary material, we show that the uni-modal version of our network performs comparably to VRCNet for partial shapes for which many complete shapes can be imagined.

\textbf{GANs and multimodality.} GAN-based techniques have also been employed for shape completion. 
Wang et al. \cite{wang2017shape} use a GAN in the form of a convolutional autoencoder to produce a coarse shape which is then sliced into images and passed to a recurrent convolutional network for refinement. 
To setup an unpaired shape completion, two autoencoders are initially trained on complete and incomplete shapes in \cite{chen2019unpaired}. 
The latent code of an incomplete shape is then mapped to a complete shape's latent code via a generator, and two discriminators are used to evaluate complete shapes and their latent codes.
The other generative method is proposed by Wu et al. \cite{wu2020multimodal} where they combine VAE and conditional GAN to achieve multimodality. Later in this paper, we show that although \cite{wu2020multimodal} is successful in point cloud completion due to mode collapse, our IMLE-based approach is more successful in producing shapes with more variety. A series of experiments have been performed to examine the effectiveness of our method, which is discussed in Section \ref{sec:experiments}.
In addition to GAN-based approaches that are capable of providing multimodality, a voxel-based generative model (GCA) is introduced by Zhang et al. \cite{zhang2021learning} that produce results by gradually completing voxels. Although we have provided a qualitative comparison with GCA in the supplementary material, it might not be directly relevant to our work as GCA is designed for voxels instead of point clouds, and also it does not use GAN in its structure while we try to provide an alternative to GAN.  
}

\section{Method}
\label{sec:method}

\noindent\textbf{Overview.} In the shape completion problem, we want to predict a complete point cloud based on a given partial point cloud with missing regions or parts. \am{In the multimodal version of this problem, a set of complete shapes is desired under only one given partial point cloud.} Formally, given an input $\mathbf{x}$ in a one-to-many problem setting, the goal is to learn $p(\mathbf{y}|\mathbf{x})$ where $\mathbf{y}$ is a prediction. Here, $\mathbf{x}$ is our partial point cloud and $\mathbf{y}$ is a predicted complete point cloud. Since there are many possible complete shapes that are consistent with $\mathbf{x}$, we want to generate non-deterministic predictions $p(\mathbf{y}|\mathbf{x})$ instead of a single prediction. The simplest way to generate non-deterministic predictions is to add a random code input $\mathbf{z}\sim \mathcal{N}(0,\mathbf{I})$ to a deterministic function $T_{\theta}$ where $\mathbf{y} \coloneqq T_{\theta}(\mathbf{x}, \mathbf{z})$. This is also known as an implicit generative model~\cite{mohamed2016learning}. One way to train this implicit generative model is to use a conditional GAN, in this context, where $T_{\theta}(\mathbf{x}, \mathbf{z})$ is often known as \emph{generator}. \am{However, the generator may fail to produce diverse outputs due to mode collapse.} \am{Therefore, we use conditional Implicit Maximum Likelihood Estimation (IMLE) to train the generative model to overcome the mode collapse issue.}

\begin{figure}[t]
    \includegraphics[width=1\linewidth]{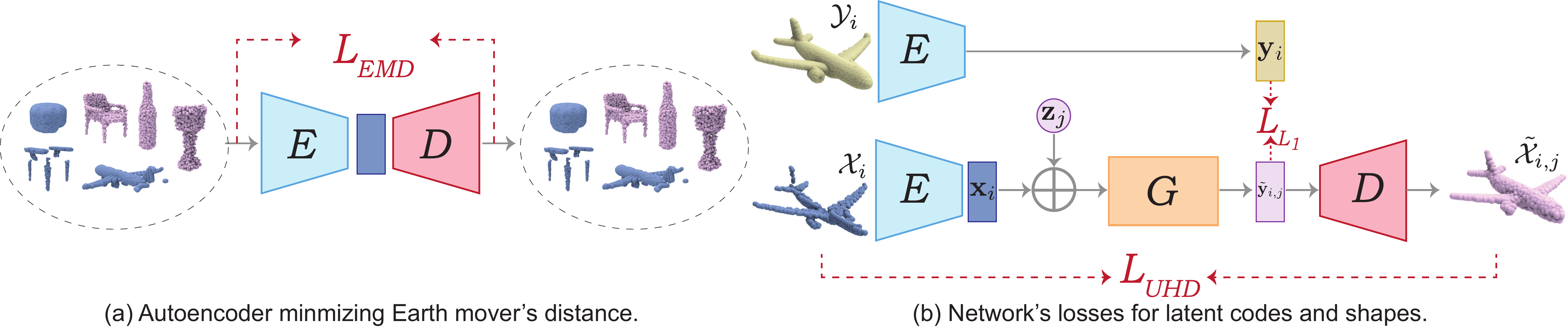}
    \caption{Autoencoder is trained on both partial (blue) and complete (purple) shapes using Earth mover's distance (a). Two losses are used on shapes and latent codes in our network (b). Losses are highlighted in red.}
    \label{fig:losses}
\end{figure}

IMLE reverses the direction of how generated samples are matched to the target data: instead of making the generated samples similar to some target data, IMLE ensures that every target data has a similar sample being generated. \am{In addition, IMLE does not require an adversarial objective which simplifies training.} If we denote the generator as $T_\theta(\cdot,\cdot)$, an input point cloud as $\mathbf{x}_i$, the corresponding ground truth as $\mathbf{y}_i$, and a latent code as $\mathbf{z}_{i,k}$, then IMLE would optimize the following equation:
\begin{equation}
    \min_{\theta}\mathbb{E}_\nv{Z}\left[\sum_{i=1}^{n}\min_{k\in \{1,\ldots,m\}}d(T_\theta(\mathbf{x}_{i},\mathbf{z}_{i,k}), \mathbf{y}_{i})\right]
\end{equation}
\am{where $\nv{Z}={\mathbf{z}_{1,1},\ldots,\mathbf{z}_{n,m} \sim \mathcal{N}(0, \mathbf{I})}$, $d(\cdot,\cdot)$ is a distance metric and, $m$ is a hyperparameter.} Algorithm~\ref{alg:cimle} shows the conditional IMLE training procedure. In our case, IMLE is conditioned on the given partial shape.

\begin{algorithm}[h]
\caption{\label{alg:cimle}Conditional IMLE Training Procedure}
\begin{algorithmic}
\Require The set of inputs $\left\{ \mathbf{x}_{i}\right\} _{i=1}^{n}$ and the set of corresponding observed outputs $\left\{ \mathbf{y}_{i}\right\} _{i=1}^{n}$
\State Initialize the parameters $\theta$ of the generator $T_\theta$
\For{$p = 1$ \textbf{to} $N$}
    \State Pick a random batch $S \subseteq \{1,\ldots,n\}$
    \For{$i \in S$}
        \State Randomly generate i.i.d. $m$ latent codes 
        \State \quad$\mathbf{z}_{1},\ldots,\mathbf{z}_{m}$
        \State $\tilde{\mathbf{y}}_{i, k} \gets T_{\theta}(\mathbf{x}_{i}, \mathbf{z}_{k})\; \forall k \in [m]$
        \State $j \gets \arg \min_{k} d(\mathbf{y}_{i}, \tilde{\mathbf{y}}_{i, k})\;\forall k \in [m]$
    \EndFor
    \For{$q = 1$ \textbf{to} $M$}
    \State Pick a random mini-batch $\tilde{S} \subseteq S$
    \State $\theta \gets \theta - \eta \nabla_{\theta}\left(\sum_{i \in \tilde{S}}d(\mathbf{y}_{i}, \tilde{\mathbf{y}}_{i, j})\right) / |\widetilde{S}|$
    \EndFor
\EndFor
\State \Return $\theta$
\end{algorithmic}
\end{algorithm}


\noindent\textbf{Shape completion network's overall structure.} 
\am{
The structure of our network is composed of an auto-encoder (i.e., encoder $E$ and decoder $D$) along with generator $G$ (see Figure \ref{fig:pipe-line}).
The auto-encoder encodes partial shape $\mathcal{X}_i$ and complete shape $\mathcal{Y}_i$ via its encoder and generates a novel and complete shape $\mathcal{\tilde{X}}_i$ via its decoder.
Generator $G$ plays the role of $T_\theta$ in Algorithm \ref{alg:cimle} by mapping the latent code $\nv{x}_i$ of the partial shape, aggregated by a noise vector $\nv{z}_j$ sampled from normal distribution, to a new code $\nv{\tilde{y}}_{i,j}$. 
$\nv{\tilde{y}}_{i,j}$ is then passed to decoder $D$ to produce a new shape $\mathcal{\tilde{X}}_{i,j}$. 
Therefore, each partial shape $\mathcal{X}_i$ can be mapped to many shapes $\mathcal{\tilde{X}}_{i,j}$.}


\noindent\textbf{Multi-modality and training.} 
\am{To train the network on the entire data, for each input $\nv{x}_i$, $m$ forward passes with different random vectors ($\nv{z}_1$,$\nv{z}_2$,...,$\nv{z}_m$) are processed. Therefore, the output from the generator is $m$ latent vectors ($\tilde{\nv{y}}_{i,1}$,$\tilde{\nv{y}}_{i,2}$,...,$\tilde{\nv{y}}_{i,m}$) where each latent vector corresponds to a different complete shape or \emph{mode}. 
As also described in Algorithm \ref{alg:cimle}, we then use a nearest neighbour algorithm to select $\tilde{\nv{y}}_{i,j}$ which is closest to the latent vector encoded from the corresponding complete shape $\nv{y}_i$. We compute an $\text{L}_1$ loss between $\tilde{\nv{y}}_{i,j}$ and $\nv{y}_i$ to bring $\tilde{\nv{y}}_{i,j}$ closer to the selected mode (see Figure \ref{fig:losses}):}
\begin{equation}
    L_{\text{L}_1} = d(\nv{y}_i,\tilde{\nv{y}}_{i,j})
\end{equation}

\am{Enforcing that every complete shape in our training data has a \emph{similar} shape in generated samples ensures that mode collapse does not occur and we have diversity in generated shapes. Finally, we decode $\tilde{\nv{y}}_{i,j}$ through decoder $D$ and produce a novel complete shape. We use Unidirectional Hausdorff Distance (UHD) loss between the given partial shape and the generated complete shape to ensure that the generated complete shape encloses the partial shape effectively. This loss can be defined as:
\begin{equation}
    L_{UHD} = \max_{x \in \mathcal{X}_i} \min_{y \in \mathcal{\tilde{X}}_{i,j}} || x - y ||,
\end{equation}
where $x$ and $y$ are single points belonging to partial and the generated complete point clouds.}

\am{Note that our auto-encoder is trained for reconstruction task on a point cloud dataset which includes both partial and complete shapes. We employed Earth Mover's Distance (EMD) loss between input and output point clouds $\nv{p}_{in}$ and $\nv{p}_{out}$ to train our auto-encoder:
\begin{equation}
    L_{EMD}(\nv{p}_{in},\nv{p}_{out}) = \min_{\xi:\nv{p}_{in} \rightarrow \nv{p}_{out}} \sum_{\alpha \in \nv{p}_{in}} ||\alpha - \xi(\alpha) ||
\end{equation}
}
\noindent\textbf{Testing.}
\am{
During testing, our model does not require any accompanied complete shape $\mathcal{Y}_i$ because of which there is no nearest neighbour selection involved. We simply pass the partial shapes along with different randomly sampled noise vectors to the network and decode all the produced latent codes into complete generated shapes.}

\section{Experiments and Results}
\label{sec:experiments}

\am{In this section, we provide various experiments and compare our work both qualitatively and quantitatively with baseline methods. In the end, we are providing some statistics about timing along with qualitative results to show how much our method respects the partial input shape. We have also provided some complementary experiments on baselines and real-world scan data in the supplementary material.}

\noindent\textbf{Baselines.}
\am{We compare the result of our point completion tasks against two multimodal methods: MSC \cite{wu2020multimodal}, and kNN-latent (a baseline introduced by \cite{wu2020multimodal}). kNN-latent is a simple baseline to look for most similar complete shapes in dataset using encoded latent vector of partial shape. MSC uses a conditional GAN architecture to generate multiple shapes. We perform quantitative and qualitative comparisons with these baselines in the following. 
In supplementary material, we provide quantitative and qualitative comparisons with other methods including uni-modal shape completion techniques such as pcl2pcl \cite{chen2019unpaired} and VRCNet \cite{pan2021variational}. We show that our method performs comparably.}

\noindent\textbf{Datasets.}
\am{We use three datasets to evaluate our work in different settings. First, we use 3Depn dataset \cite{dai2017complete} which is generated from ShapeNet using virtual scans. The dataset has varied proportions of incompleteness in the samples. We evaluate our method on three classes of this dataset - Chair, Plane, and Table. Second, we evaluate Partnet dataset \cite{chang2015shapenet} in which parts are randomly removed from a complete shape while training/testing. In Partnet, we evaluate on Chair, Table, and Lamp because of high number of samples for these classes. Finally, we use the recently released RobustPointSet \cite{taghanaki2020robustpointset} dataset for evaluating transferablity of our network. Shapes in this dataset have missing regions synthetically created on various objects.}

\begin{figure}
    \includegraphics[width=0.9\linewidth]{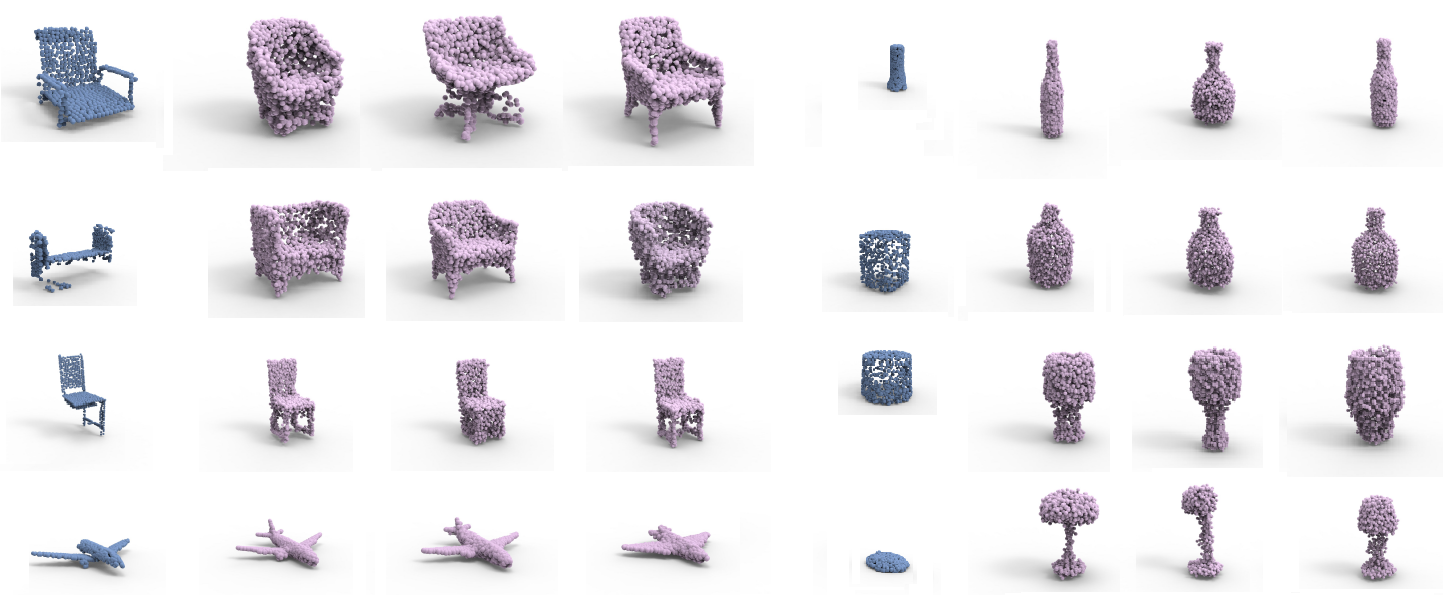}
    \caption{Our completion results for multiple object categories (e.g., Chair, Lamp, Plane and Bottle). It can be observed that the generated complete shapes are diverse for every object and respects partial shapes.}
    \label{fig:vis2}
\end{figure}

\noindent\textbf{Evaluation.}
Here we provide quantitative and qualitative results to evaluate our method against baselines. In Table \ref{tab:tmd} we compare our result using Total Mutual Difference on 3Depn (left) and Partnet (right) datasets. Total Mutual Difference (TMD) quantifies the diversity among the generated samples. This is calculated using Chamfer distance between every pair among generated samples. We show that we have significant gains in terms of diversity in generated samples in comparison with multimodal baselines. The difference can be explained by the fact that our work tends to use all modes in the dataset. Similarly, in Table \ref{tab:uhd} we compare our result using  Unidirectional Hausdorff Distance (UHD) on 3Depn (left) and Partnet (right) datasets. UHD is a measure of how well the partial shape is enclosed by the generated complete shape. We either outperform the baselines or perform relatively similarly. 
To provide qualitative results, we show our generated samples for various shape categories in Figure \ref{fig:vis2}. It can be observed that in the generated shapes, our method tends to produce various geometric and shape attributes. 

\begin{table}[t]
    \begin{minipage}{.5\textwidth}
      \centering
    
      \begin{tabular}{l|r|r|r}
\hline
Methods        & Chair                     & Plane & Table\\\hline
kNN-latent     & 2.84                      & 1.13  & 3.25                    \\
Multimodal-GAN & 2.56                      & 2.03   & 4.49                   \\
Ours           & \textbf{2.93}          & \textbf{2.31} & \textbf{4.92}\\                     \hline
\end{tabular}
    \end{minipage}
    \begin{minipage}{.5\textwidth}
      \centering
    \begin{tabular}{l|r|r|r}
\hline
Methods        & Chair                     & Lamp & Table \\\hline

kNN-latent     & 2.28                      & 4.18 & 2.85                     \\
Multimodal-GAN & 2.56                      & 3.31   & 3.88                   \\
Ours           & \textbf{2.76}                      & \textbf{5.49}    & \textbf{4.45}  
\\ \hline
\end{tabular}
\end{minipage}
\caption{Comparison between our methods and baselines in terms of Total Mutual Difference (TMD) or \textit{diversity} on two datasets - 3Depn (left) and PartNet (right). Higher TMD values indicate more diversity. TMD values in the table are multiplied by a factor of $10^2$.}
  \label{tab:tmd}
  \end{table}
\begin{table}[t]
    \begin{minipage}{.5\textwidth}
      \centering
      \begin{tabular}{l|r|r|r}
\hline
Methods        & Chair                     & Plane & Table\\\hline

kNN-latent     & 8.94                      & \textbf{9.54}    & 10.04                  \\
Multimodal-GAN & 9.14                      & 9.59          & 8.74            \\
Ours           & \textbf{8.51}                      & 9.55   & \textbf{8.52} 
\\ \hline
\end{tabular}
      
    \end{minipage}
    \begin{minipage}{.5\textwidth}
      \centering
\begin{tabular}{l|r|r|r}
\hline
Methods        & Chair                     & Plane & Table\\\hline

kNN-latent &8.58&8.47&7.61   \\
Multimodal-GAN & 6.65 &\textbf{5.40 }&5.38                  \\
Ours   &    \textbf{6.17}    &   5.58    &\textbf{5.16}             
\\ \hline
\end{tabular}
\end{minipage}
\caption{Comparison between our method and baselines in terms of Unidirectional Haussdorf Distance (UHD) or \textit{completeness} on two datasets - 3Depn (left) and PartNet (right). Lower UHD values indicate better completion. UHD values in the table are multiplied by a factor of $10^2$.}
\label{tab:uhd}
  \end{table}

\begin{figure}
    \centering
    \includegraphics[width=1\linewidth]{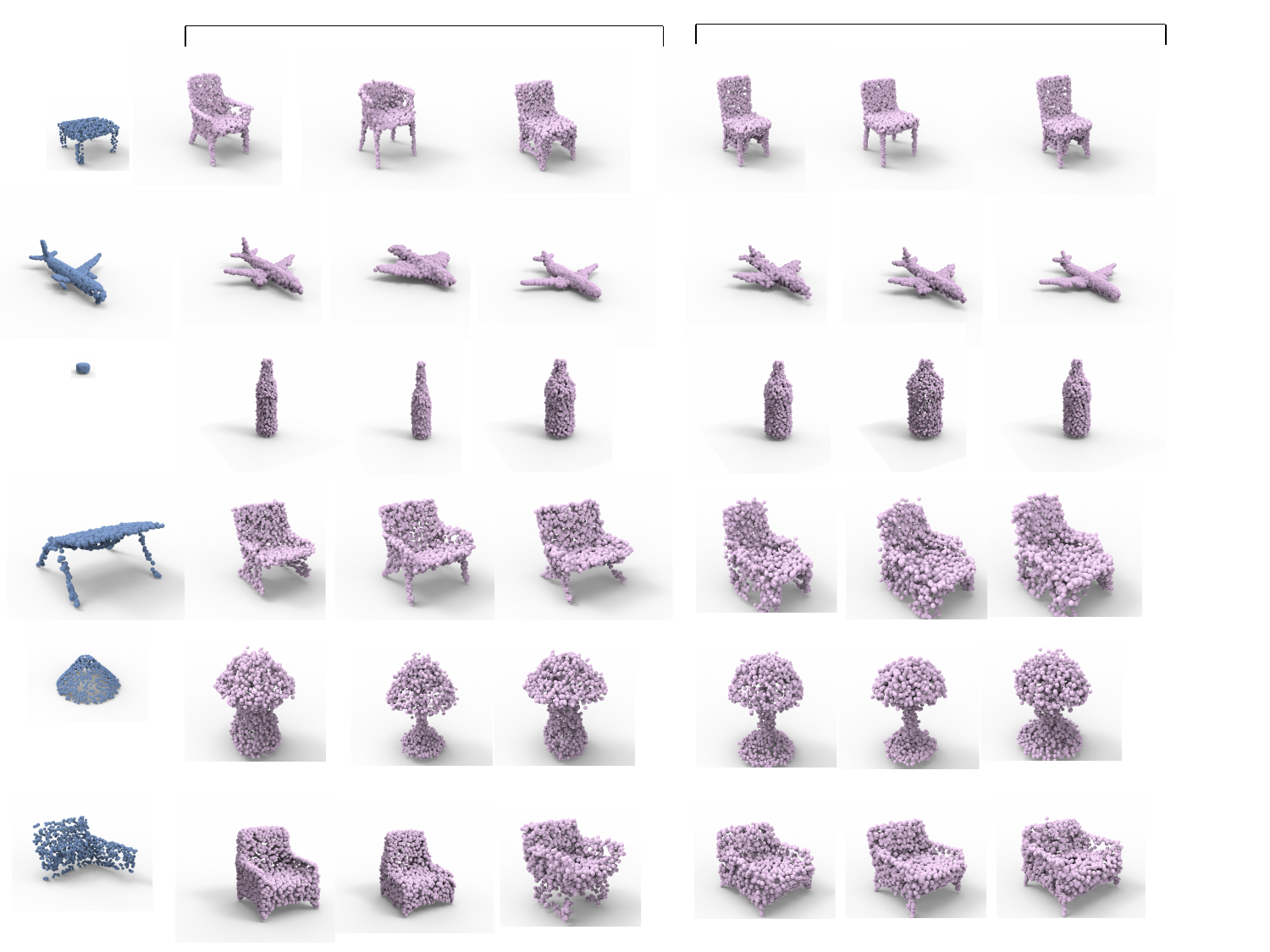}
    \caption{Qualitative Comparison of our work (left) and conditional GAN \cite{wu2020multimodal} (right). As it is apparent, our method produces more diverse objects.}
    \label{fig:vis3}
\end{figure}

\begin{figure}
 \centering
    \includegraphics[width=1\linewidth]{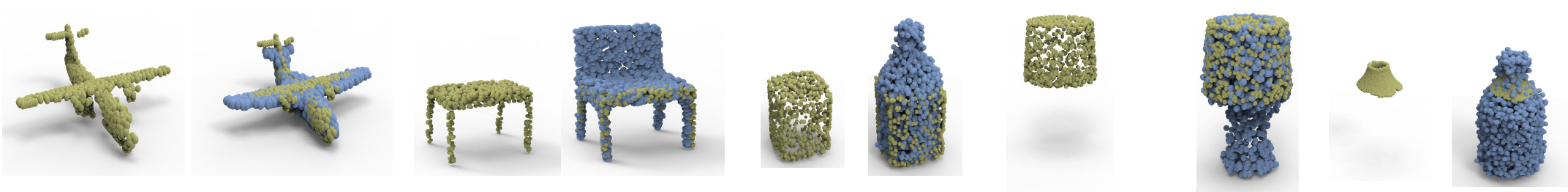}
    \caption{Partial shapes (green) and generated complete shapes (blue) are overlapped to illustrate how well our complete shape respects the partial shape.}
    \label{fig:overlap}
\end{figure}

Figure \ref{fig:vis3} contrasts our results to conditional GAN \cite{wu2020multimodal} and shows that our model has more diversity. Note that our method respects the partial input very well as apparent in Figure \ref{fig:overlap}. To make a better contrast, we have overlapped the partial input highlighted in green over the generated complete shape highlighted in blue.

\begin{figure}
    \includegraphics[width=1\linewidth]{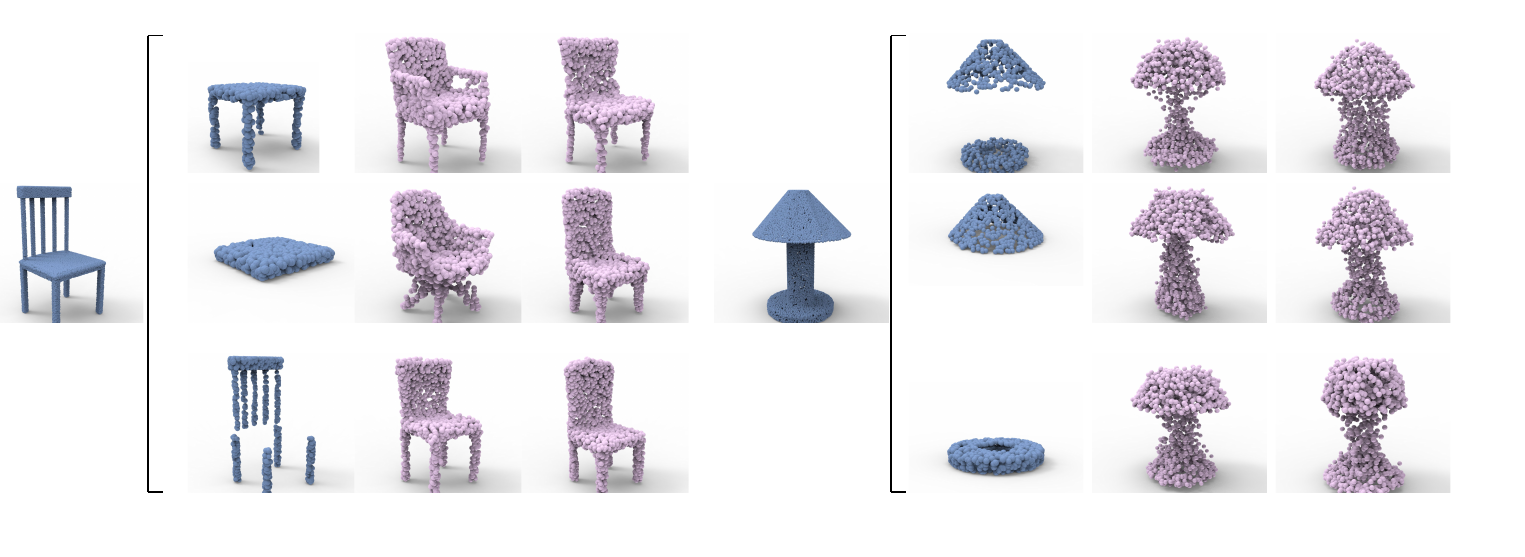}
    \caption{Three partial portions are taken from complete shapes for two categories (Chair and Lamp) and completed by our technique.}
    \label{fig:vis5}
\end{figure}

\noindent\textbf{Part-based completion.}
In Figure \ref{fig:vis5}, we demonstrate the ability of our method to generalize to unseen types of partial shapes as input. We remove a part entirely and visualize different completions generated by our method. Even though no part labels were used during training and so the model has never seen an entire part taken out, the model can generate reasonable and diverse completions. Moreover, the model can generate a diverse range of parts in place of the missing part that is functionally similar but structurally different, which suggest several interesting properties: (1) the model has learned which parts have the same function and are interchangeable, suggesting it has acquired an implicit understanding of the semantics of the part in question even though it has never seen part labels during training. (2) The model does not necessarily generate the particular shape of the part that was taken out in all completions, suggesting it did not collapse to a single mode. (3) The model is able to generate structurally diverse shapes for missing parts in different completions, suggesting that it has learned the space of structural variations for a part. Moreover, the variations can be localized to a part even though global noise is injected into the architecture.

\begin{figure}
    \centering
    \includegraphics[width=1\linewidth]{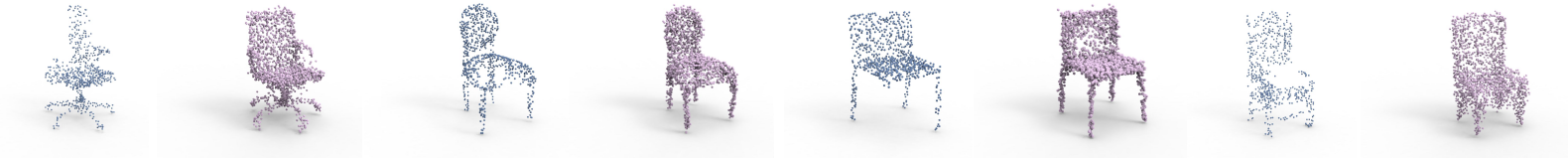}
    \caption{Generator trained on 3Depn dataset is fine-tuned on RobustPointSet dataset and results are evaluated on Chair class.}
    \label{fig:transfer}
\end{figure}

\noindent\textbf{Transfer Learning.}
To examine whether our model can be used on other datasets, possibly with more non-uniform missing parts and noises, we perform an experiment on  Robustpointset dataset \cite{taghanaki2020robustpointset}. Robustpointset contains various artifacts including missing regions, rotations, noise, etc. We use our model trained on 3Depn dataset and fine-tune it on this dataset. Our model successfully fills in or completes the missing regions and parts. In Figure \ref{fig:transfer}, we show that our model can be successfully transferred to other datasets. It is worth noting that these samples do not have much scope for a diverse generation, so our model naturally converges to the actual shape.

 \noindent\textbf{Timing Comparison.}
 Since our network does not have a discriminator, its training and inference times are lower than conditional GAN. Our method needs 13 ms per shape for training and 2 ms for inference, while conditional GAN needs 20ms per second and 2ms for inference. \cite{wu2020multimodal}. On a Geforce RTX 2080 GPU, the overall training time for conditional GAN work for the main class Chair (3Depn) is 10 hours for the autoencoder, 10 hours for the VAE and 5 hours for the generator/discriminator network. In comparison, for the same case, our work takes total 13.5 hours (10 hours for the autoencoder and 3.5 hours for the generator training).

\begin{figure}
    \centering
    \includegraphics[width=1\linewidth]{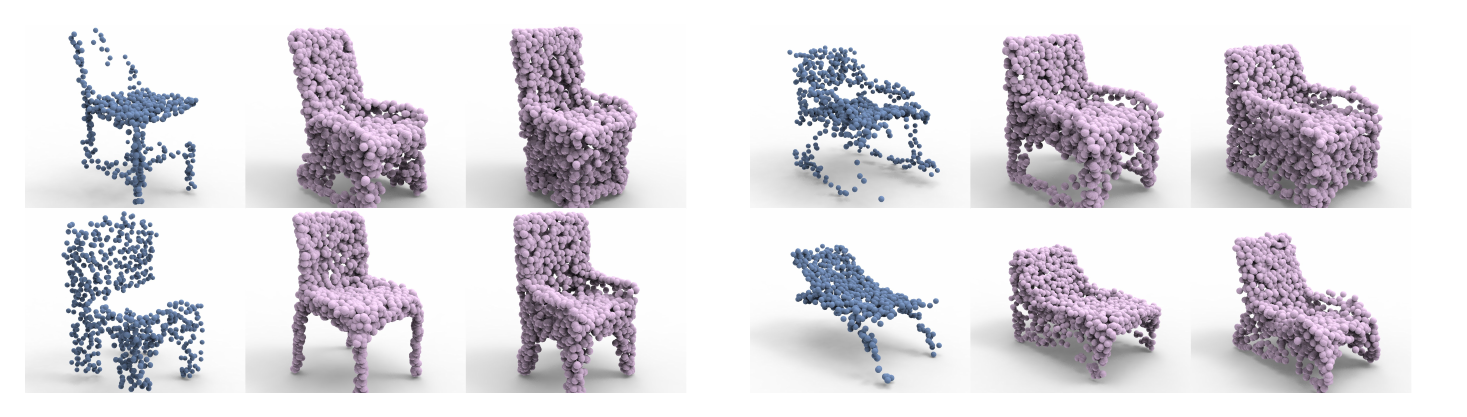}
    \caption{Noisy partial shapes are fed through the network for shape completion through our technique}
    \label{fig:noise}
\end{figure}

\noindent\textbf{Noise Tolerance.}
We perform noise tolerance analysis of our work by adding Gaussian noise to input point clouds. We perform this on 3Depn dataset for Chair category. As it can be seen in Figure \ref{fig:noise} , we are able to achieve reasonable completion despite having noise in partial shapes. We achieve $2.71$ x $10^-2$ as Total Mutual Difference value and $8.84$ x $10^-2$ as Unidirectional Hausdorff Distance value with noisy input point clouds. 

\noindent\textbf{Implementation details.}
We tune our hyperparameters through experiments to have a standard set of parameters for all classes. For the encoder and decoder model, we use 1D convolutional layers coupled with batch-normalization layers. We train this auto-encoder network for 1000 epochs with initial learning rate of $5e-4$ . Similarly, we use linear layers for the generator and train it for 500 epochs with the same initial learning rate. During testing, we set $m=10$ for the number of samples we generate for every partial shape.

\noindent\textbf{Limitations.}
Our current network cannot generate very detailed features in a multimodal manner. This is primarily due to the network's simple architecture and can be resolved by adding refinement networks.

\section{Conclusion and Future Work}
\label{sec:conclusion}
In this paper, we tackled the problem of multimodal shape completion problem, in which a given partial shape can be completed in several different forms. In fact, when a partial shape is missing a region or part, there is not a single and fixed complete shape and the result can be subject to interpretation. As a result, a method capable of producing several plausible results is desired to expand the creativity and diversity of shape completion techniques. Here, we used IMLE that tends to avoid mode collapse as opposed to conditional GAN to make sure that we can deliver diverse completed shapes. Through several experiments, we have shown that our method is superior in respecting the input partial shapes' geometry while producing more diverse results. We have also shown that without direct supervision, the network learns to respect geometric properties of shapes such as symmetry and it learns to generate missing parts in the right location with accurate geometry. The results of our shape completion are also comparable with the state-of-the-art uni-modal shape completion techniques (see supplementary material), which shows that IMLE is capable of respecting features when they are present.

Several avenues can be explored related to shape completion and also IMLE. Our shape completion only adds parts and missing pieces. However, it would be interesting to study how we can add styles and fine details in a multimodal manner. IMLE can also be used in other tasks where a one-to-many mapping is needed. An example can be single view reconstruction, where several objects can be produced according to a single input image. 

\bibliographystyle{nips}
\bibliography{ref}

\newpage
\end{document}


\section{Supplementary Material}

In this supplementary material, we provide additional quantitative and qualitative results along with comparisons with two unimodal shape completion methods (pcl2pcl \cite{chen2019unpaired} and VRCNet \cite{pan2021variational}) and a probabilistic generative model that is capable of performing multimodal shape completion (GCA \cite{zhang2021learning}). We also provide more comparison results with conditional GAN shape completion technique \cite{wu2020multimodal} and also provide qualitative results on real scan datasets.

\section{Comparison with Unimodal Methods}

Here we provide a set of comparisons with unimodal shape completion techniques to show that our method produces comparable results although it can also perform in a multimodal fashion. We compare our method with two state-of-art methods that are described in the following. 

 \textbf{pcl2pcl} \cite{chen2019unpaired} uses a GAN based network similar to \cite{wu2020multimodal} but it is not multimodal and it is unable to generate diverse results. We have chosen this GAN-based method to contrast the quality of results produced by IMLE and a GAN-based approach in a unimodal setting. 
 
 \textbf{VRCNet} \cite{pan2021variational} is based on a probabilistic modelling approach followed by local shape refinement using relational point features. This is the state-of-the-art and most recent point cloud completion method. We have selected VRCNet to be representative of point cloud completion networks that respect geometric features in the shape completion task. 

As demonstrated in Table \ref{tab:unimodal}, our results are comparable with the state-of-the-art unimodal methods since our method is capable of producing better results in Plane and Table classes while VRCNet performs better in Chair class. It should be noted that we can only compare with unimodal works on the basis of \textit{completeness} or Unidirectional Hausdorff Distance and there is no base to compare the \textit{diversity} here.

\begin{figure}[h]
    \centering
    \includegraphics[width=0.9\linewidth]{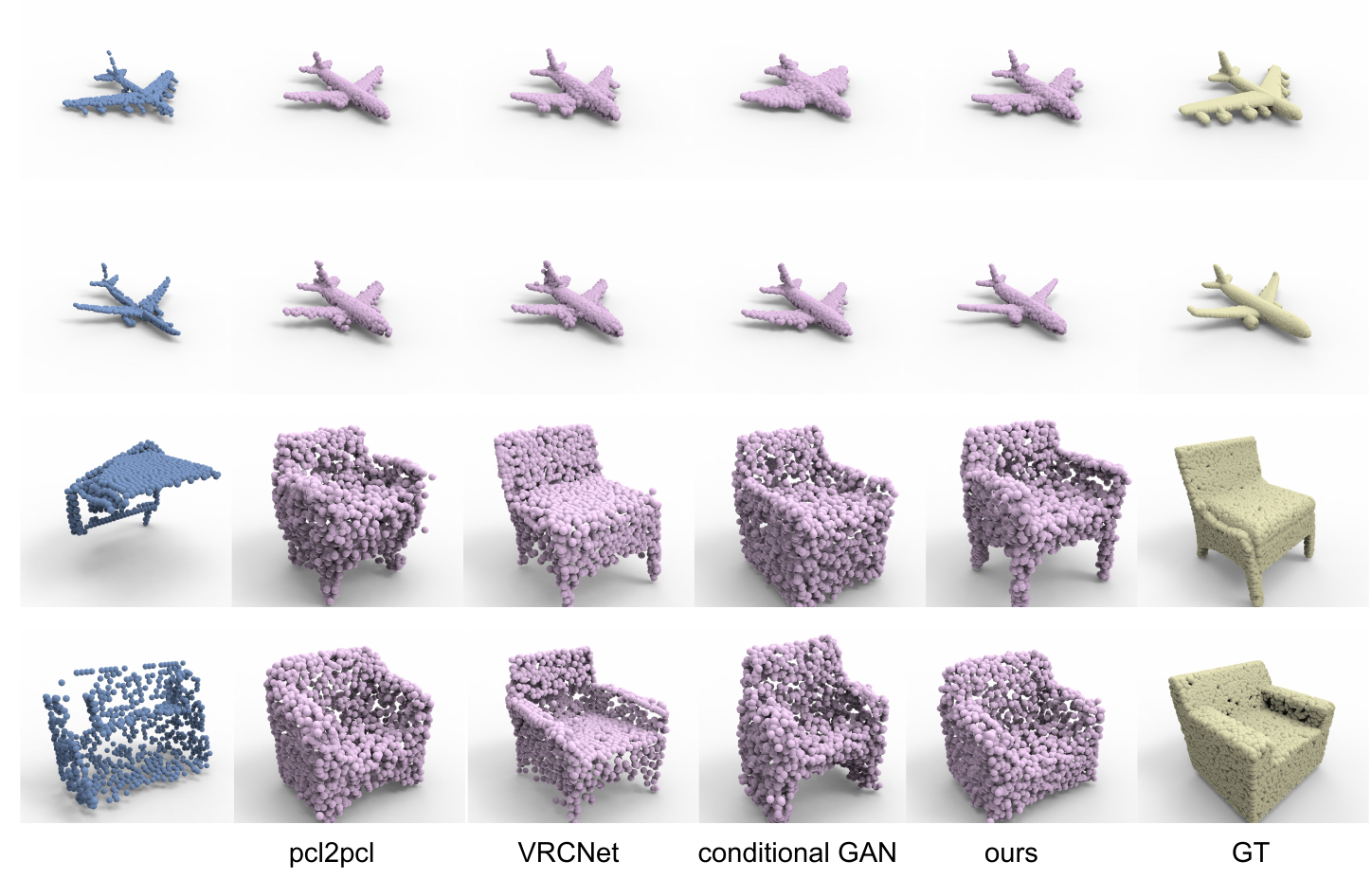}
    \caption{Qualitative comparison of pcl2pcl , VRCNet , conditional GAN and our work. }
    \label{fig:unimodal1}
\end{figure}

We also observe in Figure \ref{fig:unimodal1} that the quality of the results of our simple method is similar or superior to the state-of-the-art unimodal methods. This is despite the fact that unimodal works like VRCNet use complex networks to refine shape geometry.

\begin{table}[hbt]
    
      \centering
      \begin{tabular}{l|r|r|r}
\hline
Methods        & Chair                     & Plane & Table \\\hline

pcl2pcl        & 5.31                      & 9.71   &     9.03        \\
VRCNet    &       \textbf{5.28}               & 9.58    &    8.81              \\
Ours           & 8.51                     & \textbf{9.55}  & \textbf{8.52}
\\ \hline
\end{tabular}

\caption{Comparison between our method and unimodal methods in terms of Unidirectional Haussdorf Distance (UHD) or \textit{completeness} on 3Depn dataset.}
\label{tab:unimodal}
  \end{table}

\section{Comparison with Multimodal Methods}

\begin{figure}[h]
    \centering
    \includegraphics[width=0.9\linewidth]{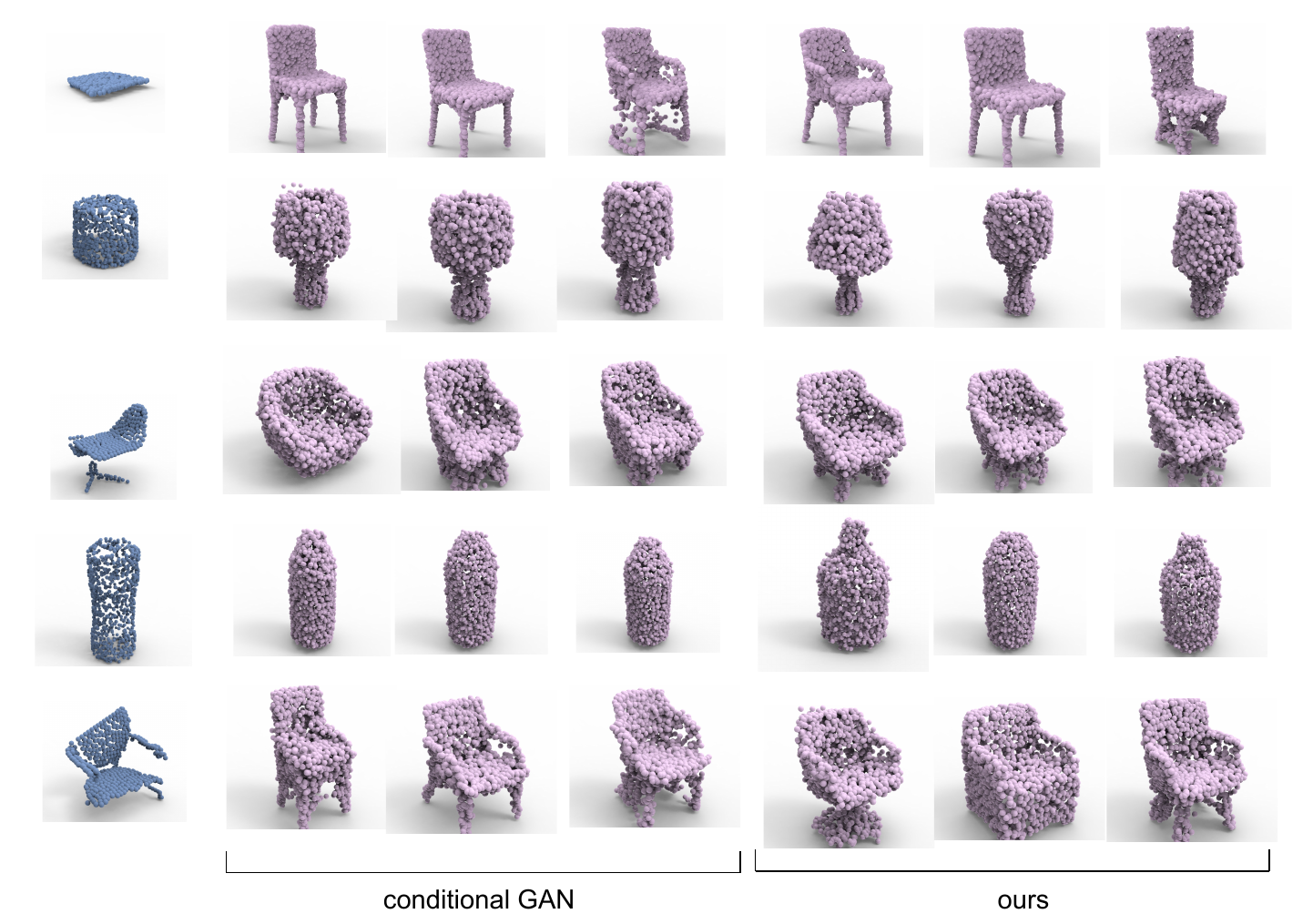}
    \caption{ Results produced by our method (right) in comparison with conditional GAN (left).}
    \label{fig:cgan2}
\end{figure}

Here, we first provide additional qualitative results against conditional GAN \cite{wu2020multimodal}. As apparent in Figure \ref{fig:cgan2}, our method can produce more diverse and clean results.

\begin{figure}
    \centering
    \includegraphics[width=1\linewidth]{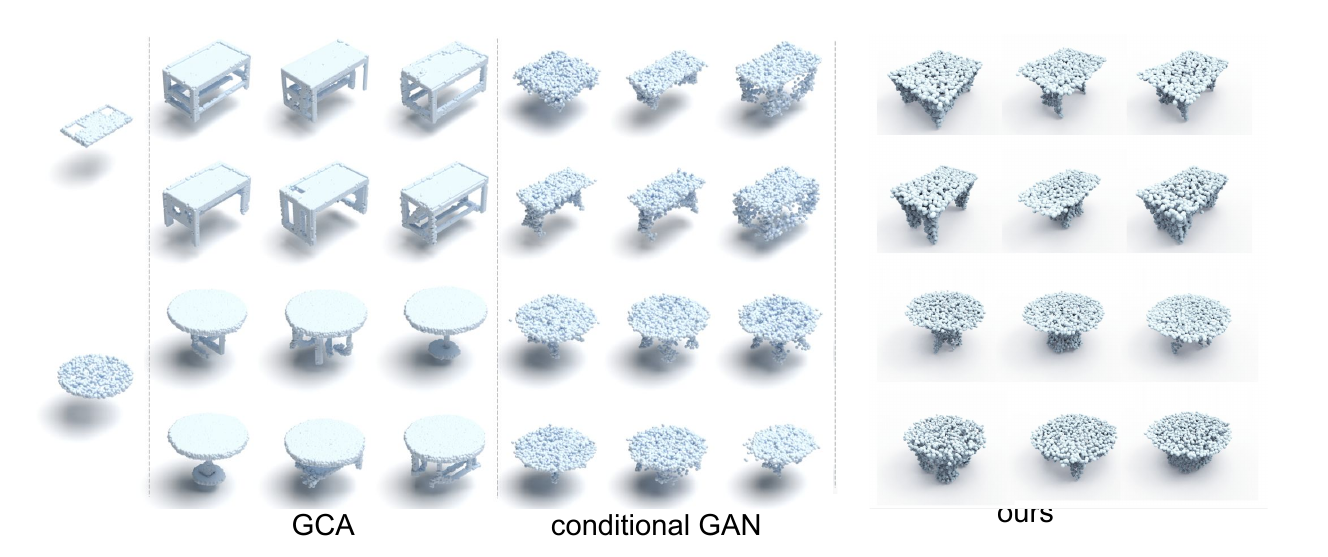}
    \caption{ Comparative results to highlight the difference between GCA \cite{zhang2021learning}, conditional GAN \cite{wu2020multimodal}, and our method. GCA results are directly taken from the paper \cite{zhang2021learning}.} 
    \label{fig:gca1}
\end{figure}

We have also provided some qualitative results with GCA \cite{zhang2021learning} that is a probabilistic model capable of performing multimodal shape completion. As illustrated in Figure \ref{fig:gca1}, both our method and GCA are capable of producing diverse results; both our method and GCA are producing various bases given the round and rectangular tops. It is also worth mentioning that due to the simplicity of our network, our inference time is only 0.02 seconds while GCA takes 0.7 seconds to complete a single inference execution since they run their network for 100 times to generate a single output. Quantitative results are not provided as the GCA's code is not publicly available yet.

\section{Real-world Scans}

\begin{figure}
    \centering
    \includegraphics[width=0.9\linewidth]{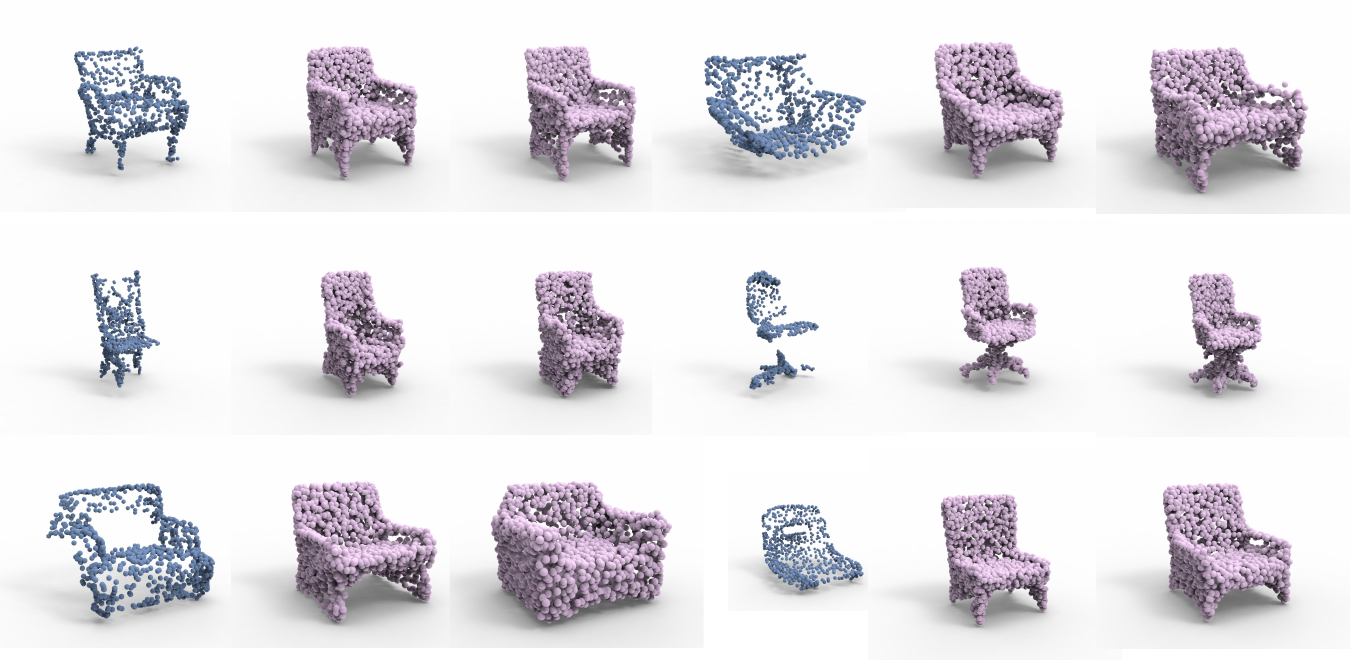}
    \caption{ Qualitative results on real-world scans from ScanNet dataset} 
    \label{fig:realScan}
\end{figure}

Here we show our results on ScanNet dataset shared by \cite{chen2019unpaired}. This dataset comprises of real-world scans of chairs. The scanned samples are more noisy and sparse as compared to other datasets which makes them a difficult dataset to evaluate on. We use the model we trained on 3Depn dataset to evaluate on real-world scans and we demonstrate successful completion with decent variety (see Figure \ref{fig:realScan}).

\clearpage
\newpage

\bibliographystyle{nips}
\bibliography{ref}